%% file: root.tex
\documentclass[letterpaper, 10 pt, conference]{ieeeconf} 

\IEEEoverridecommandlockouts     
\usepackage{amsmath,amsfonts,amssymb,amsthm,epsfig,url,array,xcolor,soul}
\usepackage{breqn}

\makeatletter
\let\NAT@parse\undefined
\makeatother

\usepackage[%
pagebackref=false,
bookmarks=true,bookmarksopen=true,
bookmarksnumbered=true,
breaklinks=true,
colorlinks=true,
anchorcolor=blue,citecolor=black,
urlcolor=blue,linkcolor=blue,filecolor=blue,
menucolor=blue,
]{hyperref}
\usepackage{subcaption}
\usepackage{cite} 

\usepackage{algorithm}
\usepackage{algpseudocode}

\newtheorem{lemma}{Lemma}

\newtheorem{theorem}{Theorem}
\newtheorem{problem}{Problem}

\newcommand{\eg}{e.g.\ }

\newcommand{\cE}{\mathcal{E}}
\newcommand{\cV}{\mathcal{V}}

\newcommand{\R}{\mathbb{R}}
\newcommand{\Zp}{\mathbb{Z}^+}

\input{preamble.tex}

\DeclareMathOperator*{\minimize}{minimize\ }
\DeclareMathOperator*{\maximize}{maximize\ }

\title{\LARGE \bf Resilience in multi-robot target tracking through reconfiguration}

\author{Ragesh K. Ramachandran, Nicole Fronda and Gaurav S. Sukhatme
	\thanks{This work was supported in part by the Army Research Laboratory as part of the Distributed and Collaborative Intelligent Systems and Technology (DCIST) Collaborative Research Alliance (CRA). The authors are with the Department of Computer Science, University of Southern California, Los Angeles, CA 90089, USA  {\tt\small rageshku| nfronda|gaurav@usc.edu}}
}

\begin{document}
\maketitle
\thispagestyle{empty}
\pagestyle{empty}

\begin{abstract}

We address the problem of maintaining resource availability in a networked multi-robot system performing distributed target tracking. In our model, robots are equipped with sensing and computational resources enabling them to track a target's position using a Distributed Kalman Filter (DKF).  We use the trace of each robot's sensor measurement noise covariance matrix as a measure of sensing quality. When a robot's sensing quality deteriorates, the system’s communication graph is modified by adding edges such that the robot with deteriorating sensor quality may share information with other robots to improve the team's target tracking ability. This computation is performed centrally and is designed to work without a large change in the number of active communication links. We propose two mixed integer semi-definite programming formulations (an `agent-centric' strategy and a `team-centric' strategy) to achieve this goal. We implement both formulations and a greedy strategy in simulation and show that the team-centric strategy outperforms the agent-centric and greedy strategies.

\end{abstract}

\section{Introduction and Related work}
\label{sec: intro}

Target tracking using a robot team is of interest in many civilian and military applications. Consequently, this has led to a burgeoning interest in studying target tracking problems using robots~\cite{hausman2015cooperative,olfati2011collaborative,williams2015,dames2017detecting}. Even though each robot has a limited view of the environment, multiple robots could perform the target tracking task collaboratively by exchanging information among themselves. A necessary condition for this to succeed is the existence of an underlying communication network. Collaborative distributed tracking avoids the need for a central data fusion center which combines the data collected on the robots to estimate the state of the target. A common approach for multi-robot target tracking in a distributed manner is through a distributed Kalman filter\cite{olfati2011collaborative}. We follow this method in this paper. 

\begin{figure}
		\centering
		\includegraphics[width=\linewidth]{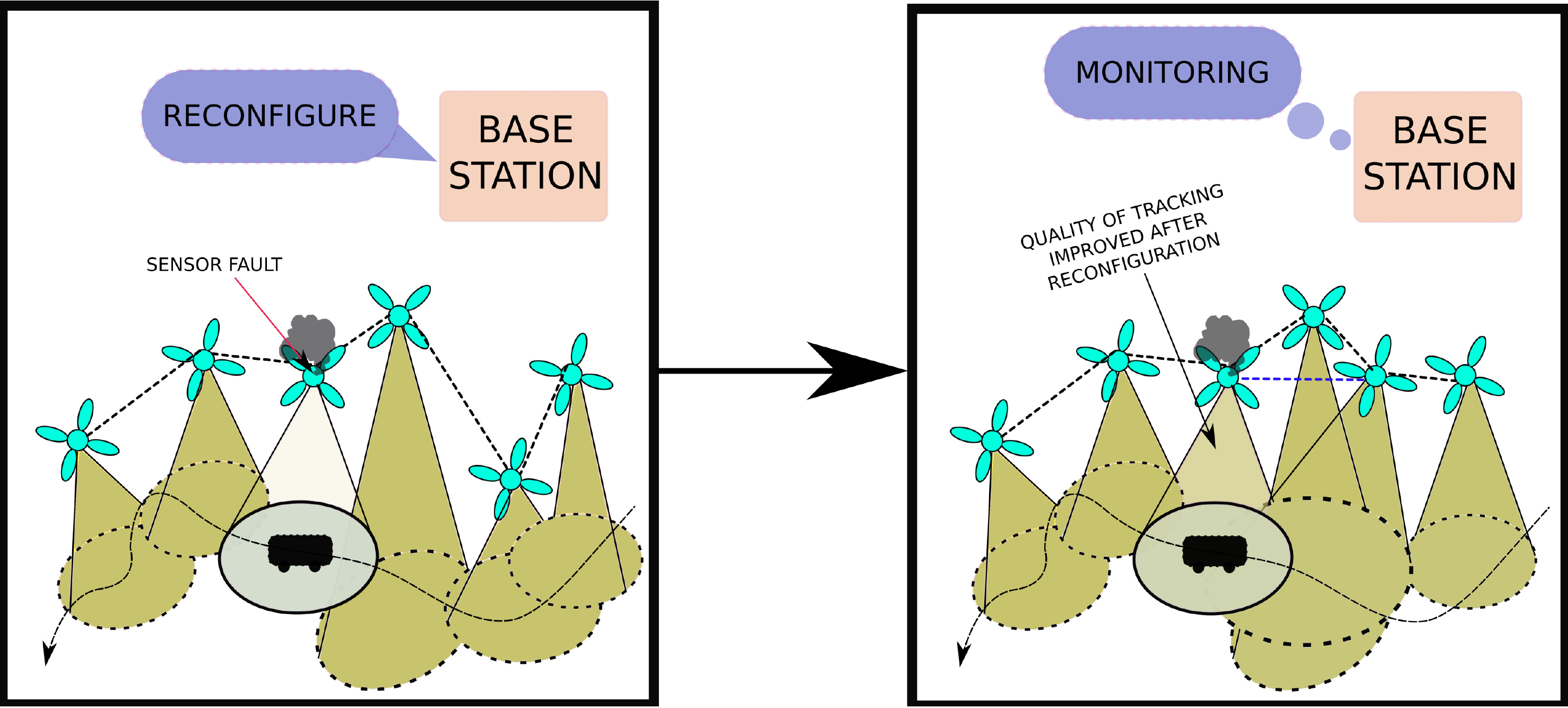}
		\caption{\small{A setting for resilient target tracking.}}
		\vspace{-9mm}
		\label{fig:failure eg}       
	\end{figure}

We envision a scenario in which a team of robots tracks and estimates the state of a target while running a distributed Kalman filter on board. For simplicity we assume there is a single sensor (\eg a camera) on each robot that is used to obtain target information. The robot team is monitored by a base station which intervenes in the team's activities only if a robot in the team experiences a deterioration in sensing quality. We use the trace of the sensor's measurement noise covariance matrix as the measure of sensing quality. Our focus is on the strategies to mitigate the effect of a sensor quality deterioration on a robot (and hence on the team's tracking performance) by appropriately modifying the topology of the underlying communication network and target state measurement fusion weights (Figure \autoref{fig:failure eg}). In a control theoretic sense, the state estimation error of the DKF depends upon the observability of the underlying communication network \cite{liu2017kalman}. Understanding and quantifying the effect of network topology on a network's observability properties \cite{Pasqualetti2014,leitold2017controllability,Ramachandran2017} is well studied. These findings form the basis of our proposed strategy to employ communication network topology reconfiguration to mitigate the effect of sensor quality deterioration in target tracking performance.  Although, we restrict ourselves to single target tracking in this paper, our technique can be applied to multi-target tracking assuming that robots can identify and discern individual targets. 

The framework discussed here extends the abstract framework described in \cite{ramachandran2019resilience} by adapting it to multi-robot target tracking. Similar to \cite{ramachandran2019resilience}, we propose a two-stage strategy to alleviate the impact of a robot's sensor quality deterioration on the team's performance. Our framework is in two stages. The first stage updates the team's communication network topology such that the new topology is adjacent to the original. It also computes an optimal set of new measurement fusion weights. Specifically, we optimizes a monotonic function of the trace of the covariance matrix of the target state estimation error. The second stage generates a set of coordinates implementing communication network topology while maximizing the coverage of the robots over the target's domain.  The framework considered here only accounts for sensor quality deterioration and not for the situation where the sensor fails completely. However, a near-complete sensor failure can be modeled as a sensor with very high measurement noise covariance. We also assume that the sensor measurement noise is always characterized by a zero mean probability distribution. Finally, we assume that robots can estimate their sensor quality. Since there exist techniques for sensor fault detection \cite{sharma2010sensor} and degradation estimation \cite{arosh2015fitness,jiang2006sensor}, this assumption does not impose undue restrictions on the applicability of our method. 

Stage one of our method uses \textit{mixed integer semi-definite programs} (MISDPs) to formulate and solve the problem of fabricating a communication graph topology and associated information fusion weights that improves the tracking performance of the team. We consider two MISDPs: \textit{agent-centric configuration generation} (\textbf{ACCG}) and \textit{team-centric configuration generation} (\textbf{TCCG}). The first MISDP(\textbf{ACCG}) maximizes the trace of the inverse of \textit{target state estimation error covariance matrix} (\textbf{TSEECM}) associated with the robot that experienced the sensor quality degradation. The second one (team-centric) minimizes the average of the trace of \textbf{TSEECM} over all the robots. Our procedure in stage two solves an optimization problem to compute the coordinates of the robots that maximize the coverage of the robots over the target's configuration space under communication constraints dictated by the graph topology computed in stage one. Although, resilience in multi-robot systems is well studied  \cite{Bonilla2017}, the idea of resilience through reconfiguration during task execution is recent. This paper introduces the idea of resilience by reconfiguration into multi-robot target tracking.


\section{Notation and Preliminaries} 
For any positive integer $z \in \Zp$, $[z]$ denotes the set $\{1,2, \cdots, z\}$. $\|\cdot\|$ denotes the standard Euclidean 2-norm and the induced 2-norm for vectors and matrices. $\|\mathbf{M}\|_F$ is the Frobenius norm of the matrix $\mathbf{M} \in \R^{m_1 \times m_2}$. $Tr(\mathbf{M})$ is the trace of matrix $\mathbf{M}$. $\mathbf{1}^{m_1}$ and $\mathbf{0}^{m_1}$ are the vector of ones and zeros with appropriate dimensions.  We drop the superscripts whenever the dimensions of the vectors or matrices are clear from the context. We use $\mathbf{\Bar{0}}_i^{n}$ to denote a vector of zeros with one as its $i^{th}$ entry. For any vector $\mathbf{v}$, $Diag(\mathbf{v})$ is the matrix with $\mathbf{v}$ along its diagonal. Conversely, $diag(\mathbf{M})$ yields the vector containing the diagonal elements of the matrix $\mathbf{M}$. $\mathbf{M}^{\text{T}}$ or $(\mathbf{M})^{\text{T}}$ is the transpose of $\mathbf{M}$. $\mathcal{S}^m_+$ denotes the space of $m \times m$ symmetric positive semi-definite matrices. A weighted undirected graph with non negative edge weights $\mathcal{G}$ is defined using the triplet $(\mathcal{V},\mathcal{E} \subseteq \mathcal{V} \times \mathcal{V}, \mathbf{A} \in \R^{|\mathcal{V}| \times |\mathcal{V}|}_{\geq 0})$, where $\mathbf{A}$ is the weighted adjacency matrix of the graph. 
$\overline{\cE} = (\cV \times \cV) \setminus \cE$ is the edge complement of $\mathcal{G}$. A matrix $\mathbf{M}$ is doubly stochastic if its rows and columns sum to unity \cite{Horn:1985:MA:5509}. Mathematically, $\mathbf{M} \mathbf{1}^n = \mathbf{1}^n$ and $\mathbf{1}^{\text{T}}\mathbf{M}  = \mathbf{1}^{\text{T}}$. One is an eigenvalue of any doubly stochastic matrix and all other eigenvalues of a doubly stochastic matrix have a magnitude less than one (\cite{FB-LNS}, Lemma 2.9). 

\section{Problem Formulation}\label{sec:problem}

We consider a team of \textit{n} robots whose labels belong to $\{1, 2, \cdots, n\}$.  The team is tasked with tracking a moving target of interest for a time period of $T$ epochs. It is assumed that the moving target is confined to a compact Euclidean space $\mathcal{D}$.  We refer to the robot team that tracks this moving target as the \textit{tracker team} and the robots as  \textit{trackers}. The robot with label $i$ is indicated as $\rho^i$. Let $X^i$ denotes the triplet position vector $[x_i, y_i, z_i] \in \R^3 $ of $\rho^i$, then the set $\{X_{[n]}\}$ contains the positions of all trackers. Also, $X_{target}$ represents the position vector of the target. We assume that the trackers are equipped with localization capabilities which enable them to localize with reasonable accuracy in the environment.  We consider the scenario where the tracker team performs distributed target tracking, thus trackers need to communicate among themselves during the tracking task.

Let the dynamic undirected graph $\mathcal{G}[k] = (\mathcal{V}, \mathcal{E}[k], \mathbf{A}_u[k])$ model the communication network of the tracker team at the $k^{th}$ time step($k \in [T]$). Note that we use time step, time and epoch interchangeably in this paper. The node set $\mathcal{V}$ is isomorphic to the tracker team label set $[n]$. An edge $(i,j)$ is included in the edge set $\mathcal{E}[k]$ if $\rho^i$ communicates with $\rho^j$ at time $k$. We denote the communication range of trackers as $d_{mc}>0$ . The neighbor set of node $i$ in $ \mathcal{G}[k]$ is defined as $\mathcal{N}_{(i)}[k] = \{ j \in \mathcal{V}:  (i,j) \in \mathcal{E}[k] \}$. The interaction between nodes in a graph can also be represented using an unweighted adjacency matrix. The adjacency matrix of $\mathcal{G}[k]$ is denoted by $\mathbf{A}_u[k]$.

\subsection{Distributed Kalman filter for tracking}
\label{subsec: distrib Kalman filter}

We describe the distributed Kalman filter computation employed by the robots to collectively track a target maneuvering in an environment according to some known dynamics. The main advantage of employing a distributed (as opposed to a centralized) Kalman filter is that it eliminates the need for a central data fusion center \cite{olfati2007dkf,olfati2011collaborative,liu2017kalman,morbidi2011active}. We follow the formulation in \cite{battistelli2012consensus,liu2017kalman}. The target dynamics are described by the standard linear state space equation 
\vspace{-0.15in}
\begin{align}
    \label{eqn:tracking agent dynamics}
    \mathbf{x}_{k+1} = \mathbf{F}_k\mathbf{x}_{k} + \mathbf{G}_k\mathbf{u}_k + \mathbf{w}_k,
\end{align}
where $\mathbf{x}_{k} \in \R^{s_a}$ and  $\mathbf{u}_k \in \R^{u_a}$ are the state and the input vectors of the target respectively. $\mathbf{F}_k \in \R^{s_a \times s_a}$ and $\mathbf{G}_k \in \R^{s_a \times u_a}$ are the state transition matrix and input matrix of appropriate dimensions respectively. $\mathbf{w}_k \in \R^{s_a}$ is the zero mean normally distributed random vector with the covariance matrix $\mathbf{Q}_k \in \R^{s_a \times s_a}$ ($\mathbf{w}_k \sim \mathcal{N}(\mathbf{0}, \mathbf{Q}_k)$).  Each tracker can obtain measurements about the state of the  target. The sensing region of $\rho^i$ is a disc of radius $d^i_{sen}> 0$ centered around the robot.  The measurement model of $\rho^i$ in its sensing region is $\mathbf{z}^i_k = \mathbf{H}^i_k \mathbf{x}_{k} + \mathbf{v}^i_k$
where $\mathbf{z}^i_k \in \R^{m_i}$ and $\mathbf{H}^i_k \in \R^{s_a \times m_i}$ are the measurement vector and output matrix of $\rho^i$, respectively. Also, $\mathbf{v}^i_k \in \R^{m_i}$ is a zero mean Gaussian sensing noise vector with a covariance matrix $\mathbf{R}^i_k \in \R^{m_i \times m_i}$ modeling the sensor noise characteristics of $\rho^i$. 

The distributed Kalman filter algorithm consists primarily of two steps: 1) \textit{individual update} and 2) \textit{consensus update}. In the individual update step, the robots that receive a measurement estimate the state of the tracked agent using the standard Kalman filter equations. Mathematically, if $\rho^i$ received a measurement $\mathbf{z}^i_k$ then it makes an estimate about the state of the tracked target using the following equations. 

\textit{Prediction:}
\vspace{-0.15in}
\begin{align}
    \label{eqn: state prediction}
    \mathbf{\Hat{x}}^i_{k|k-1} &=  \mathbf{F}_{k}\mathbf{\Hat{x}}^i_{k-1} + \mathbf{G}_{k}\mathbf{u}_{k-1} \\
    \label{eqn: covariance prediction}
    \mathbf{P}^i_{k|k-1} &= \mathbf{F}_k \mathbf{P}^i_{k-1} (\mathbf{F}_k)^{\text{T}} + \mathbf{Q}_k
\end{align}

\textit{Local innovation or measurement update:}
\vspace{-0.1in}
\begin{align}
    \label{eqn: kalman gain}
    \mathbf{K}^i_k &= \mathbf{P}^i_{k|k-1}(\mathbf{H}^i_k)^{\text{T}}\left(\mathbf{H}^i_k \mathbf{P}^i_{k|k-1} (\mathbf{H}^i_k)^{\text{T}} + \mathbf{R}^i_k \right)^{-1} \\
    \label{eqn: state innovation}
    \mathbf{\Hat{x}}^i_{k,0} &=  \mathbf{\Hat{x}}^i_{k|k-1} + \mathbf{K}^i_k(\mathbf{z}^i_k - \mathbf{H}^i_k \mathbf{\Hat{x}}^i_{k|k-1}) \\
    \label{eqn: covariance innovation}
    \mathbf{P}^i_{k,0} &=  \mathbf{P}^i_{k|k-1} - \mathbf{K}^i_k \mathbf{H}^i_k \mathbf{P}^i_{k|k-1}.
\end{align}

The local target state estimate $\mathbf{\Hat{x}}^i_{k,0}$ of $\rho^i$ is fused with the local target state estimate of other trackers in the tracker team in a distributed fashion via a consensus protocol. This distributed information fusion is performed in the consensus step of the distributed Kalman filter. Each tracker in the tracker team computes the information vector $\mathbf{q}^i_{k}(0) = (\mathbf{P}^i_{k,0})^{-1} \mathbf{\Hat{x}}^i_{k,0}$ and the associated information matrix $\mathbf{\Omega}^i_{k}(0) = (\mathbf{P}^i_{k,0})^{-1}$ prior to the consensus step. Once these quantities are computed, each robot in the tracker team exchanges information with its neighbors. Ultimately this results in consensus on a refined state estimate of the tracked target. The robots exchange information as follows: 
\begin{align}
    \label{eqn: consnss omega updte}
    \mathbf{\Omega}^i_{k}(l+1) &= \sum_{j \in \mathcal{N}_{(i)}[k] \cup i} [\mathbf{\Bar{A}}[k]]_{i,j} \mathbf{\Omega}^j_{k}(l) \\
    \label{eqn: consnss q updte}
    \mathbf{q}^i_{k}(l+1) &= \sum_{j \in \mathcal{N}_{(i)}[k] \cup i} [\mathbf{\Bar{A}}[k]]_{i,j} \mathbf{q}^j_{k}(l),
\end{align}

where $[\mathbf{\Bar{A}}[k]]_{i,j}$ is the $(i,j)$ entry of a doubly stochastic matrix $\mathbf{\Bar{A}}[k]$ with the same structure as the unweighted adjacency matrix ($\mathbf{A}_u[k]$) of $\mathcal{G}[k]$ except for the diagonal elements. Specifically, $\mathbf{\Bar{A}}[k]$ is non-zero along its diagonal and its off-diagonal elements are non-zero if and only if the corresponding elements of $\mathbf{A}_u[k]$ are unity. In theory, information vectors and matrices of the robots converge to a common quantity only when $l$ tends to infinity. However, it is known that this consensus protocol has an exponential rate of convergence \cite{FB-LNS}. As a result, a reasonable level of consensus on the information vector and matrix can be achieved by propagating \autoref{eqn: consnss q updte} and \autoref{eqn: consnss omega updte}  for a sufficient number of consensus steps $L$. We note that, the consensus update is assumed to happen at a much faster time scale compared to \autoref{eqn:tracking agent dynamics}. Since the $L$ steps consensus process results in a fusion of all the measurements obtained by the team, the posteriori estimate of the target's state vector $\mathbf{\Hat{x}}^i_k$ and the posteriori estimation error covariance matrix $\mathbf{P}^i_{k}$ can be determined using  $ \mathbf{\Hat{x}}^i_{k}= (\mathbf{\Omega}^i_{k}(L))^{-1} \mathbf{q}^i_{k}(L)$ and $\mathbf{P}^i_{k} = (\mathbf{\Omega}^i_{k}(L))^{-1}$ respectively.

\subsection{Tracking under Sensor Quality Deterioration}

As mentioned in \autoref{sec: intro}, we consider the problem of mitigating the effect of sensor quality deterioration on target tracking performance through appropriate reconfiguration of the tracker team. Next we give a precise definition of tracker team reconfiguration and sensor quality deterioration. 

We term the tuple $(\mathcal{G}[k], \mathbf{\Bar{A}}[k])$ as the \textit{configuration} of the tracker team at the $k^{th}$ time step and denote it by $\mathcal{C}[k]$. $\mathbf{\Bar{A}}[k]$ is a doubly stochastic matrix whose elements are used to perform the information fusion computations outlined in \autoref{eqn: consnss q updte} and \autoref{eqn: consnss omega updte} for the consensus step. During tracking operation for a time period $T$, let $n_f$ detrimental events occur independently to random trackers in the tracker team. We assume that each event results in some sensor quality deterioration. At time $k$, we say that $\rho^i$'s sensor quality is deteriorated if the trace of the measurement noise covariance matrix associated with the sensor $\mathbf{R}^i_k$ has increased with respect $Tr(\mathbf{R}^i_{k-1})$. In other words, if $Tr(\mathbf{R}^i_k)$ $>$ $Tr(\mathbf{R}^i_{k-1})$, then $\rho^i$ sensor quality deteriorated at time $k$. Recall that, we assume the sensor is unbiased even after its quality deteriorates. 
Consider a sequence set $\mathcal{F} = [f_1, f_2, \cdots, f_p, \cdots, f_{n_f}]$, where $f_p$ indicate the time step when the $p^{th}$ sensor fault occurred. We specify that $\mathcal{C}[f_p-1]$ is the configuration of the tracker team before the $p^{th}$ detrimental event occurred. We now formally define the problems studied in this paper. The first problem (\autoref{prob: configuration generation}) deals with reconfiguration of the tracker team such that target tracking performance is optimal in some reasonable sense. The second problem addresses the issue of realizing the graph topology in 3D space while maximizing the tracker team's coverage over the $\mathcal{D}$. 

\begin{problem}
		\label{prob: configuration generation}
		\textbf{Configuration generation or reconfiguration:}
		Given that $\rho^i$ experienced sensor quality deterioration at some time $f_p$, $\mathbf{R}^i_{f^+_p}$ is the sensor noise covariance matrix immediately after the deterioration event and $\mathcal{C}[f_p-1]$ is the tracker configuration prior to the event determine a new configuration $\mathcal{C}[f_p]$ such that,
		\begin{enumerate}
		    \item $\mathcal{G}[f_p]$ is a connected graph, 
		    \item $\|\mathbf{A}_u[f_p] - \mathbf{A}_u[f_p-1] \|_F \leq 2\times n_e$, where $n_e \in \Zp$ is the number of edges that may be modified in $\mathcal{G}[f_p-1]$ to obtain $\mathcal{G}[f_p]$ and
		    \item tracking performance is optimized.
		\end{enumerate}
\end{problem}

\begin{problem}
        \label{prob: formation synthesis}
        \textbf{Formation synthesis:} Given a tracker team configuration $\mathcal{C}[f_p]$, generate coordinates that best realize the given configuration and maximize the tracker team's coverage over $\mathcal{D}$, subject to constraints. We defer the  details of this problem to \autoref{subsec: formation syn}.
\end{problem}

Graph connectivity is an essential requirement for any distributed computation over a network and thus is enforced in \autoref{prob: configuration generation} \cite{FB-LNS}. The second condition enables the user to control the communication load on the generated configuration by tuning the parameter $n_e$. Finally, 
the third condition ensures good tracking performance. 

\begin{figure}
		\centering
		\includegraphics[width=0.8\linewidth]{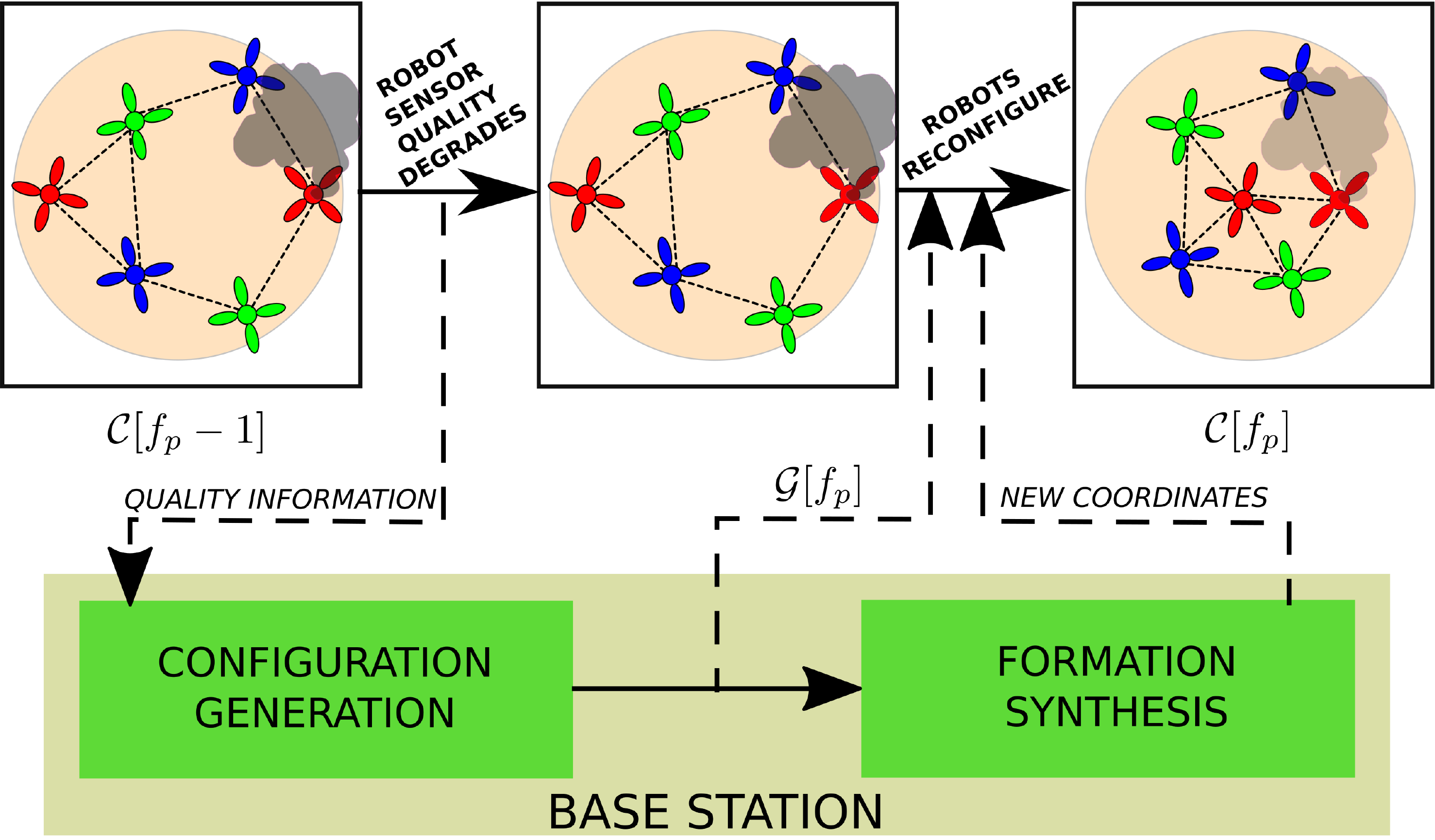}
		\caption{\small{Basic outline of our approach. When a robot experiences sensor quality degradation, \emph{configuration generation} selects edges to modify the communication graph. Then, \emph{formation synthesis} assigns robots to physical locations that support the desired graph topology. }}
		\label{fig:schematic}       
		\vspace{-8mm}
\end{figure}

\section{Methodology}
\label{sec:Methodology}

We now describe our strategies for solving \autoref{prob: configuration generation} and \autoref{prob: formation synthesis}. A base station monitors the activities of the tracker team. Only when sensor deterioration occurs, the base station directs the tracker team to reconfigure to a new formation which it computes using the available information. Following the framework in our previous work \cite{ramachandran2019resilience}, the base station uses a two-step procedure to compute a new configuration and generate a set of robot coordinates which realize the computed configuration. As in \cite{ramachandran2019resilience}, we refer to these steps as \textit{configuration generation} and \textit{formation synthesis}. The base station decision making process is shown in \autoref{fig:schematic}. The configuration generation and formation synthesis steps are solutions to \autoref{prob: configuration generation} and \autoref{prob: formation synthesis}. 

\subsection{Configuration Generation}
\label{subsec: config gen}

We propose two strategies for tackling the configuration generation problem. As mentioned earlier, we allude to these strategies as \textit{agent-centric configuration generation} (\textbf{ACCG}) and \textit{team-centric configuration generation} (\textbf{TCCG}). Both these strategies amount to solving a mixed integer semi-definite program. Network topology design problems are often addressed by posing them as MISDPs \cite{rafiee2010optimal,xue2014distributed}. In the agent-centric configuration method, the idea is to find a new configuration such that the trace of the information matrix (inverse of TSEECM) associated with the deteriorated tracker in the tracker team  is maximized. The team-centric method minimizes the average of the trace of TSEECMs associated with all the trackers. Since the optimization over the covariance matrices resulting from $L>1$ consensus steps is a very hard problem to solve, we use the TSEECM obtained after one step consensus to formulate and solve the MISDP for both strategies. Before we describe the details of the MISDP formulations, we state a theorem (see~\cite{} for a proof, omitted here for brevity) which serves as the basis for the connectivity inequality constraint in MISDPs. 

\begin{theorem}
\label{theorem:graph connectivity}
If a graph containing self loops at every node is equipped with a weighted adjacency matrix $\mathbf{A}$ which is doubly stochastic then any graph isomorphic to this graph with or without self loops is connected if and only if 
\begin{align}
    \frac{1}{n} \mathbf{1} \mathbf{1}^{\text{T}}  + \mathbf{I}  \succ \mathbf{A}
\end{align}
\end{theorem}


\subsubsection{\textbf{ACCG}}
\label{subsubsec: agent centric}
The following MISDP models our agent-centric configuration generation approach: 

\begin{align}
	\label{eqn: agent MISDP obj}
	\maximize_{\substack{\mathbf{A} \in \mathcal{S}^n_+,\ \mu \in \R_{> 0}, \\ \boldsymbol{\Pi} \in \{0,1\}^{n\times n}}} \quad &   (\mathbf{\Bar{0}}_i^{n})^{\text{T}} \mathbf{A}   \begin{bmatrix}
Tr(\mathbf{\Omega}^1_{f_p}(0))\\ 
Tr(\mathbf{\Omega}^2_{f_p}(0))\\
\vdots\\
Tr(\mathbf{\Omega}^n_{f_p}(0)) 
\end{bmatrix}\\
	\label{eqn:cons:unity sum}
	\text{subject to} ~~ & \mathbf{A}\cdot\mathbf{1}^n = \mathbf{1}^n\\
	\label{eqn:cons:connectivity}
	~~ &\frac{1}{n} \mathbf{1} \mathbf{1}^{\text{T}}  + (1 - \mu) \mathbf{I}  \succeq \mathbf{A}, \mu  \ll  1 \\
	\label{eqn:cons:diag binary}
	~~ & diag(\boldsymbol{\Pi}) = \mathbf{1}^n\ \\
	\label{eqn:cons:binary sym}
	~~ & \boldsymbol{\Pi} = \boldsymbol{\Pi}^T\\
	\label{eqn:cons:adj sym}
	~~ & \mathbf{A} = \mathbf{A}^T\\
	\label{eqn:cons:adj diag}
	~~ & [\mathbf{A}]_{i,i} > 0 ~ \forall\ i \in [n]\\
	\label{eqn:cons:adj off diag min}
	~~ & [\mathbf{A}]_{i,j} \geq 0  \forall~(i, j) \in [n]^2,~i \neq j \\
	\label{eqn:cons:adj off diag max}
	~~ & [\mathbf{A}]_{i,j} \leq  \boldsymbol{\Pi}_{i,j} \forall ~(i, j) \in [n]^2,~i \neq j\\
	\label{eqn:cons:topology near}
	~~ & \|\boldsymbol{\Pi} - \mathbf{\Bar{A}}[f_p-1] \|_F^2 \leq 2 \times n_e.
\end{align}

The decision variables $\mathbf{A}$ and $\mathbf{\Pi}$ model the doubly stochastic matrix used for the consensus protocol and the adjacency matrix of the generate configuration respectively. \hyperref[eqn:cons:unity sum]{Constraint~\ref{eqn:cons:unity sum}} and  \hyperref[eqn:cons:adj sym]{Constraint~\ref{eqn:cons:adj sym}}
 to \hyperref[eqn:cons:adj off diag max]{Constraint~\ref{eqn:cons:adj off diag max}} ensures that $\mathbf{A}$ is a doubly stochastic matrix that is structurally equivalent to $\mathbf{\Pi}$. In the light of \autoref{theorem:graph connectivity}, \hyperref[eqn:cons:connectivity]{Constraint~\ref{eqn:cons:connectivity}} enforces the generated configuration to possess a connected graph. Finally, \hyperref[eqn:cons:topology near]{Constraint~\ref{eqn:cons:topology near}} encodes the second condition in \autoref{prob: configuration generation} into the MISDP. If $i$ represents the label of the robot whose sensor quality deteriorated at $f_p$, then with some simple algebraic manipulation it is easy to see that \autoref{eqn: agent MISDP obj} is equal to $Tr(\mathbf{\Omega}^i_{f_p}(1))$ or $Tr((\mathbf{P}^i_{f_p}(1))^{-1})$.

\subsubsection{\textbf{TCCG}}
\label{subsubsec: team centric}

Consider the following MISDP formulation encoding the team-centric configuration generation strategy which minimizes $ \frac{1}{n} \sum_i^{n} \text{Trace}(\mathbf{P}^i_{f_p}(1)))$, where $\mathbf{P}^i_{f_p}(1)) = (\mathbf{\Omega}^i_{f_p}(1))^{-1}$.

\vspace{-0.3in}
\begin{align}
	\label{eqn: team MISDP obj}
	\minimize_{\substack{\mathbf{A} \in \mathcal{S}^n_+,\ \mu \in \R_{> 0}, \\ \boldsymbol{\Pi} \in \{0,1\}^{n\times n}\mathbf{\Bar{P}},\mathbf{\Bar{\Delta}} \in \mathcal{S}^{n \times s_a}_+, \\
	 \mathbf{\Delta_1}, \mathbf{\Delta_2},\cdots, \mathbf{\Delta_n} \in \mathcal{S}^{s_a}_+}}\  
	 & 
	 \frac{1}{n} Tr(\mathbf{\Bar{P}})~ 
	\text{subject to} ~~ 
	\begin{bmatrix}
	\mathbf{\Bar{P}} & \mathbf{I} \\
	\mathbf{I} &  \mathbf{\Bar{\Delta}}
	\end{bmatrix} ~ \succeq 0 \\
	~~ \mathbf{A} \otimes \mathbf{I}\begin{bmatrix}
\mathbf{\Omega}^1_{f_p}(0)\\ 
\mathbf{\Omega}^1_{f_p}(0)\\
\vdots\\
\mathbf{\Omega}^1_{f_p}(0) 
\end{bmatrix} &= \begin{bmatrix} 
\mathbf{\Delta_1}\\ 
\mathbf{\Delta_2}\\
\vdots\\
\mathbf{\Delta_n} 
\end{bmatrix}
~\label{eqn:cons:compact version of consensus}
\end{align}



Where $\mathbf{\Bar{\Delta}}$ is a block diagonal matrix with $\{\mathbf{\Delta}_1, \mathbf{\Delta}_2, \cdots, \mathbf{\Delta}_n\}$ along its diagonal and $\mathbf{A} \otimes \mathbf{I}$ results in the \textit{Kronecker product} \cite{Horn:1985:MA:5509} between $\mathbf{A}$ and the identity matrix of the equal size. \hyperref[eqn:cons:compact version of consensus]{Constraint~\ref{eqn:cons:compact version of consensus}} is essentially \autoref{eqn: consnss omega updte} for $L=1$ written compactly as a single equation for the whole tracker team. Therefore, $\mathbf{\Delta}_i$ should match the information matrix $\mathbf{\Omega}^i_{f_p}(1)$. The following lemma (see~\cite{} for a proof, omitted here for brevity) proves that minimizing \autoref{eqn: team MISDP obj} minimizes $ \frac{1}{n} \sum_i^{n} \text{Trace}(\mathbf{P}^i_{f_p}(1)))$. 

\begin{lemma}
\label{lemma: trace bound}
The  $\frac{1}{n} Tr(\Bar{\mathbf{P}})$ is an upper bound on $\frac{1}{n} \sum_i^{n} \text{Trace}(\mathbf{P}^i_{f_p}(1)))$
\end{lemma}


\subsection{Formation Synthesis}
\label{subsec: formation syn}

We now describe a procedure to assign a physical location to each robot which maximizes the non-overlapping coverage of the space.  We also impose constraints so that connected robot pairs remain within communication distance $d_{mc}$ of each other, and that the distance between all robot pairs exceeds $d_s$ to ensure that no two robots collide. An additional constraint is added to ensure that robots which were tracking the target at $f_p$ have the target within their sensing range $d_{sen}$ after reconfiguration.


We define coverage of the space as the total area covered by all robot sensing regions minus the overlapping area covered by two or more robot sensing regions. This produces the following constrained optimization problem:

\vspace{-0.2in}
\begin{align}
    \label{obj-coverage}
    \hspace{-0.1in}
	\max_{\{X_{|n|}\}} \pi \sum_{i \in \cV} \left(({d^i_{sen}})^2 \hspace{-0.05in} - \hspace{-0.05in} \sum_{j \in \cV \neq i}\frac{(2d^i_{sen} - \|X_i - X_j\|)^2}{2} \right)
\end{align}
	
	\begin{align}
	\label{con-collision}
	\vspace{-0.3in}
	\text{subject to} \quad & d_s \leq \|X_i - X_j\| \ \leq d_{mc} & & \forall\ (i, j) \in \cE \\
	\label{con-nonneighbor}
	& d_{s} \leq \|X_i - X_j\| & & \forall\ (i, j) \in \overline{\cE} \\
	\label{con-box}
	& B^{\min} \leq X_i \leq B^{\max} & & \forall\ i \in V \\
	\label{dist2target}
	& \|X_i - X_{target}\| \ \leq d^i_{sen} & & \forall\ i \in \mathcal{V}_{f_p}
	\end{align}
	where $d^i_{sen}$ is the radius of the circular field of vision of tracker, $\rho^i$,
	and $B^{\min},\ B^{\max} \in \R^3$ are the minimum and maximum extents of an axis-aligned bounding box, with the operator $\leq$ applied elementwise in~\autoref{con-box}. $\mathcal{V}_{f_p} \subseteq \cV$ is the labels of the subset of trackers tracking the target at $f_p$.

\begin{figure}
    \centering
    \includegraphics[width=0.9\linewidth]{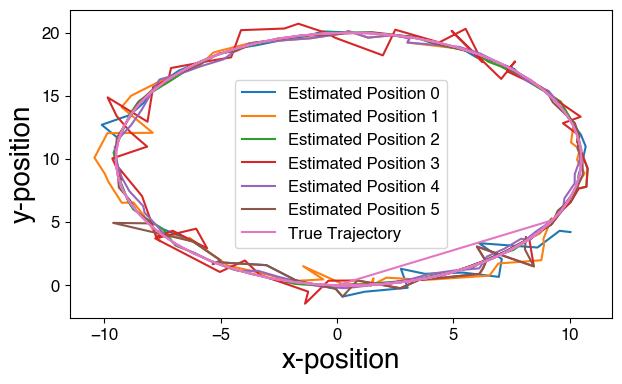}
    \caption{\small{Comparison of the target agent's true trajectory and its estimate by each robot in a tracking team of six, labeled 0-5. 
    }}
    \label{fig:sim trajectory}
\end{figure}

We solve the formation synthesis optimization problem \autoref{obj-coverage} - \autoref{dist2target} following the approach described in \cite{ramachandran2019resilience} by simply updating the objective to \autoref{obj-coverage} and incorporating the new constraint \autoref{dist2target}.

\begin{figure*}[t]
	\centering
	\begin{tabular}{c|c|c}
		\subcaptionbox{Before Second Event}{\includegraphics[width=.32\linewidth]{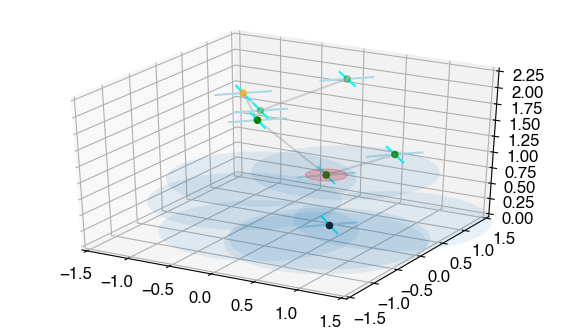}}
		&
	    \subcaptionbox{Before Fourth Event}{\includegraphics[width=.32\linewidth]{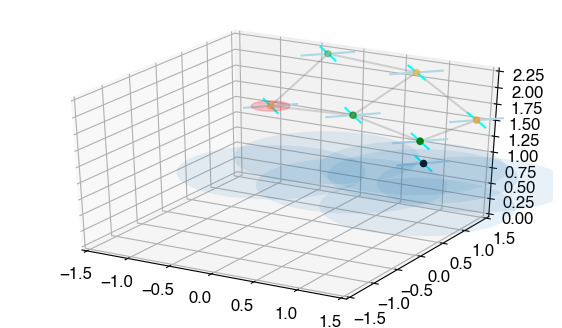}} 
		&
		\subcaptionbox{Before Sixth Event}{\includegraphics[width=.32\linewidth]{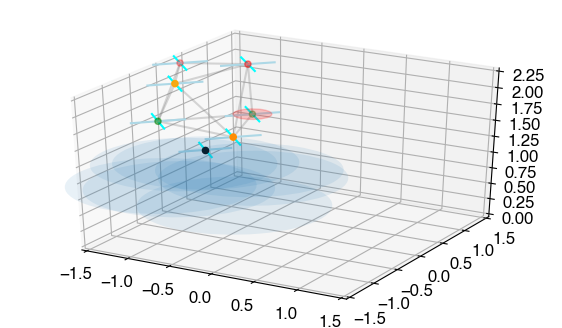}}
		\\
		\subcaptionbox{After Second Event}{\includegraphics[width=.32\linewidth]{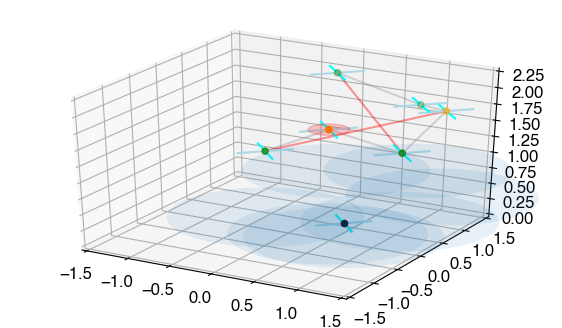}}
		&
	    \subcaptionbox{After Fourth Event}{\includegraphics[width=.32\linewidth]{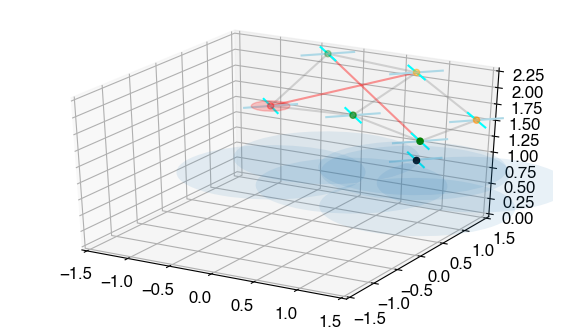}} 
		&
		\subcaptionbox{After Sixth Event}{\includegraphics[width=.32\linewidth]{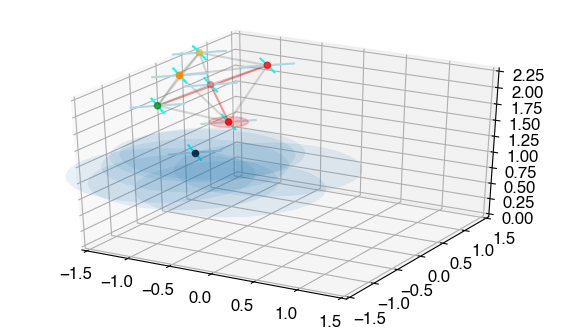}}
	\end{tabular}
	\caption{\small{Screenshots of a simulation in which a six quadrotor team tracks another quadrotor flying below them. The tracking team's sensing area is colored in light blue. The figures in the top row depict the formation of the quadrotors before the occurrence of a sensor deterioration event. The corresponding figures on the bottom portray the formation after 1) sensor deterioration is detected, 2) a new communication edge is chosen, and 3) the robots move to their new locations. The quadrotor which experienced sensor deterioration is enveloped using a filled red circle. The quadrotor colors vary from green to red indicating good to poor sensor quality.  For visualization purposes, we set $n_e = 2$ for this simulation to illustrate multiple edges added for deterioration events. We also set $d_{s}=0.5$, $d_{mc}=1.0$, and $d^i_{sen} = 1.0 \forall\ i \in [n]$, with a bounding box of $x \in [-1.5, 1.5]$, $y \in [-1.5, 1.5]$, and $z \in [1.6, 2.25]$.}}  
	\label{fig:sim failures}
\end{figure*}

\begin{figure*}[t]
	\centering
	\begin{tabular}{cc|cc}
		{\includegraphics[width=.46\linewidth]{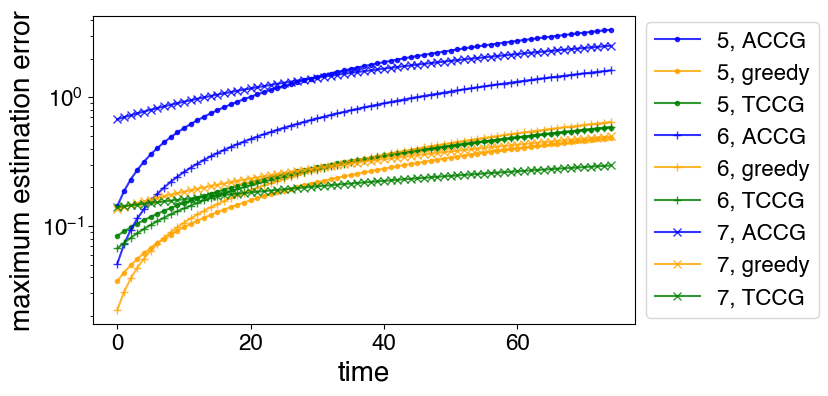}}
		&
	    {\includegraphics[width=.46\linewidth]{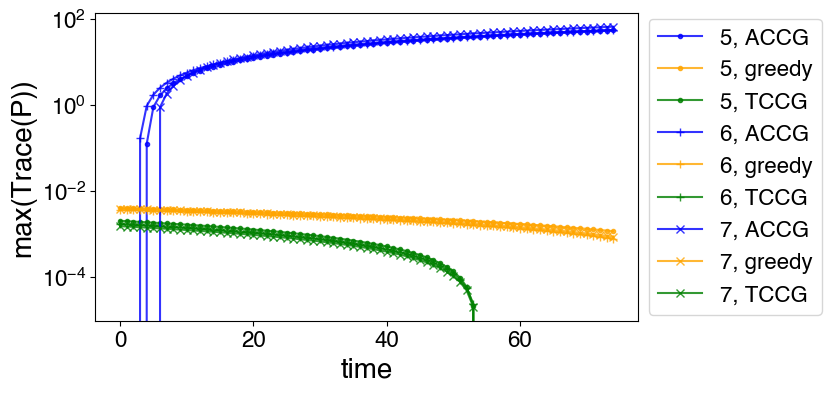}}
	\end{tabular}
	\caption{\small{The comparison (log scale) of \textit{maximum estimation error} and \textit{maximum $Tr(\mathbf{P})$} for all simulations and all team sizes between \textbf{ACCG} (blue), \textbf{TCCG} (green), and the \textit{greedy} approaches (orange). \textbf{TCCG} performs better over time in terms of estimation error for teams of $n = 6$ and $n = 7$.  \textbf{TCCG} also performs best in terms of the trace of the covariance matrix for all team sizes.}}
	\label{fig:sim validation}
\end{figure*}


\begin{figure*}
	\centering
	\begin{tabular}{ccc}
	    {\includegraphics[width=.46\linewidth]{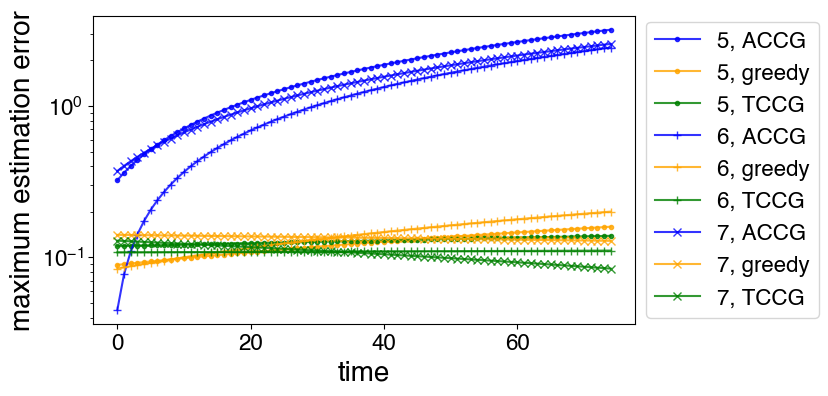}}
		&
	    {\includegraphics[width=.46\linewidth]{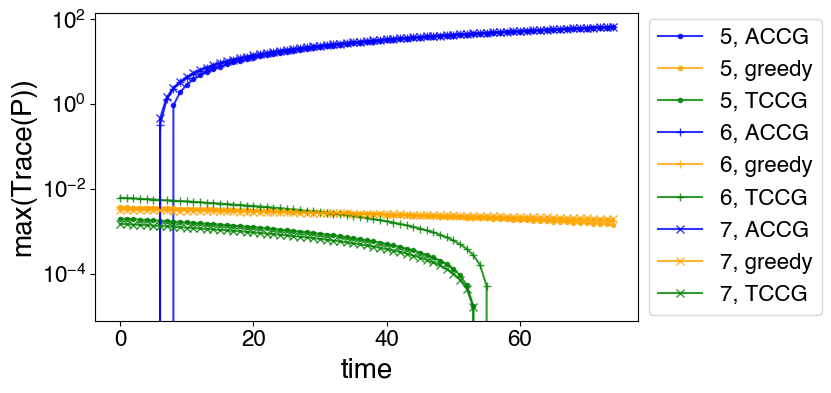}}
	\end{tabular}
	\caption{\small{A comparison of \textit{maximum estimation error} and \textit{maximum} $Tr(\mathbf{P})$ between the three different strategies for simulations in which only a single node experienced sensor deterioration. Here, \textbf{TCCG} performs best for all simulated team sizes.}}
	\label{fig:sim validation single node failure}
\end{figure*}



\section{Simulation}
\label{sec:simulation}

To validate our approach, we conducted multiple simulation experiments of a robot team tracking a target agent following a circular trajectory. We modeled the target trajectory as a Dubins car \cite{LaValle:2006:PA:1213331}. In our simulation, the target's state vector $\mathbf{x}_{k} = [x_k, y_k, \theta_k, v_k]$ describes its position, orientation and velocity at each time step \textit{k}. The parameters for the linearized discrete dynamics (\autoref{eqn:tracking agent dynamics}) of the target are



\begin{align}
    \mathbf{F}_k = \begin{bmatrix}
    1 & 0 & 0 & 0 \\
    0 & 1 & 0 & 0 \\
    0 & 0 & 1 & 0 \\
    0 & 0 & 0 & 0
    \end{bmatrix},
    \mathbf{G}_k = \begin{bmatrix}
    \cos{\theta_k}dt & 0 \\
    \sin{\theta_k}dt & 0 \\
    0 & dt \\
    1 & 0
    \end{bmatrix}
\end{align}

and input vector $\mathbf{u}_k = [\Tilde{v}_k, \Tilde{\theta}_k]$.  We held $\Tilde{v}_k$ and $\Tilde{\theta}_k$ constant throughout the simulation. This resulted in the target following a circular path.   

For the distributed Kalman filter, we initiate the same $\mathbf{H}^i_k$ and $\mathbf{R}^i_k$ for each robot in the tracker team.  We used $L=15$ for the consensus step.  Parameters chosen for the configuration generation and formation synthesis problems were 
$n_e = 1$, $d_{s}=5$, $d_{mc}=10$, and $d^i_{sen} = 20  \forall\ i \in [n]$, with a bounding box of $x \in [-100, 100]$, $y \in [-100, 100]$, and $z \in [-100, 100]$.

The target dynamics and distributed Kalman filter were implemented in Python.  For the agent-centric approach, the MISDP problem was solved using Python with PICOS as the optimization problem modeling interface and MOSEK as the semi-definite programming solver.  For the team-centric approach, the MISDP problem was solved in MATLAB with YALMIP as the modeling interface.  In both approaches, the simulated annealing technique for formation synthesis was implemented in Python.

To simulate deteriorating sensor quality for a robot \textit{i}, we modified its covariance matrix $\mathbf{R}^i_k$ by adding a random positive definite matrix.  We generated various deterioration event sequences for robot teams of $n \in \{5, 6, 7\}$  where a random robot was chosen at every \textit{f} time step of the simulation to experience sensor deterioration.  \autoref{fig:sim trajectory} shows an overhead view of a single simulation trial. The true trajectory of the target and the estimated trajectory of the target by each tracker in the team are shown.
\autoref{fig:sim failures} shows screenshots from a separate trial showing different deterioration instances before and after reconfiguration.  A video of this trial is included in the supplementary material.

For comparison of the \textbf{ACCG} and \textbf{TCCG}, we simulated 20 deterioration event sequences for each size robot team.

We also compared both \textbf{ACCG} and \textbf{TCCG} against a \textit{greedy} strategy in which a single edge is added connecting the robot whose sensing quality has deteriorated to the robot with lowest $Tr(\mathbf{P}_{f_p-1})$ at the time of the deterioration event. Each trial was initialized with a line graph.

To quantify performance, we plotted the \textit{maximum $Tr(\mathbf{P})$} and \textit{maximum estimation error} of the robot team for each strategy at each time step over all trials.   In our plots $\mathbf{P}$ is the covariance matrix of a tracker after completing $15$ consensus steps of the distributed Kalman filter, or $\mathbf{P}^i_{f_p}(15))$.  This differs from $\mathbf{P}^i_{f_p}(1))$, which we optimized in the MISDP formulations. \autoref{fig:sim validation} shows the results of these simulations.  

While the \textit{greedy} strategy appears to perform best in the early steps of the simulation, \textbf{TCCG} outperforms both \textbf{ACCG} and \textit{greedy} strategies for larger teams as the simulation progresses and sensor quality deteriorates.  Interestingly, \textbf{TCCG} performed poorly even compared to the simple \textit{greedy} approach for all team sizes.

To simulate a worst case scenario, we additionally ran simulations in which the same robot experienced sensor deterioration at each time step.  Under these conditions, \textbf{TCCG} outperforms \textbf{ACCG} and the greedy strategy at most time steps, as seen in \autoref{fig:sim validation single node failure}, for all simulated team sizes.


\section{Conclusion}
\label{sec:conclusion}
We describe a novel method that enables a team of robots engaged in tracking a target to reconfigure themselves in response to deterioration in the sensing quality in one of the team members. The reconfigured team alleviates the effect of sensor quality deterioration on team performance. We proposed two methods (both based on optimizing MISDPs) to generate the new configuration. The methods (agent-centric) and the other method (team-centric) were validated in simulation and compared to each other. They were also compared to a greedy strategy. The team-centric approach outperforms both the agent-centric and greedy approaches. We are currently working on validating our approach on our multi-robot testbed~\cite{crazyswarm}. We are also extending our approach to handle multi-robot multi-target tracking scenarios where the number of targets are unknown.

\input{root.bbl}
\end{document}

%% file: preamble.tex
\makeatletter
\@ifpackageloaded{hyperref}{%
	\@ifpackageloaded{amsmath}{%
		\newcommand{\AMShreffix}[1]{%
			\expandafter\let\csname AMShreffix#1\expandafter\endcsname%
			\csname #1\endcsname%
			\expandafter\renewcommand\csname #1\endcsname{%
				\@hyper@itemfalse\csname AMShreffix#1\endcsname}}
		\AtBeginDocument{%
			\AMShreffix{equation}
			\AMShreffix{align}
			\AMShreffix{alignat}
			\AMShreffix{flalign}
			\AMShreffix{gather}
			\AMShreffix{multline}
	}}{}}{}
\makeatother

\def\figurename{Fig.}

%% file: root.bbl
\begin{thebibliography}{10}
\providecommand{\url}[1]{#1}
\csname url@rmstyle\endcsname
\providecommand{\newblock}{\relax}
\providecommand{\bibinfo}[2]{#2}
\providecommand\BIBentrySTDinterwordspacing{\spaceskip=0pt\relax}
\providecommand\BIBentryALTinterwordstretchfactor{4}
\providecommand\BIBentryALTinterwordspacing{\spaceskip=\fontdimen2\font plus
\BIBentryALTinterwordstretchfactor\fontdimen3\font minus
  \fontdimen4\font\relax}
\providecommand\BIBforeignlanguage[2]{{%
\expandafter\ifx\csname l@#1\endcsname\relax
\typeout{** WARNING: IEEEtran.bst: No hyphenation pattern has been}%
\typeout{** loaded for the language `#1'. Using the pattern for}%
\typeout{** the default language instead.}%
\else
\language=\csname l@#1\endcsname
\fi
#2}}

\bibitem{hausman2015cooperative}
K.~Hausman, J.~M{\"u}ller, A.~Hariharan, N.~Ayanian, and G.~S. Sukhatme,
  ``Cooperative multi-robot control for target tracking with onboard sensing,''
  \emph{The International Journal of Robotics Research}, vol.~34, no.~13, pp.
  1660--1677, 2015.

\bibitem{olfati2011collaborative}
R.~Olfati-Saber and P.~Jalalkamali, ``Collaborative target tracking using
  distributed kalman filtering on mobile sensor networks,'' in
  \emph{Proceedings of the 2011 American Control Conference}.\hskip 1em plus
  0.5em minus 0.4em\relax IEEE, 2011, pp. 1100--1105.

\bibitem{williams2015}
R.~K. {Williams} and G.~S. {Sukhatme}, ``Observability in topology-constrained
  multi-robot target tracking,'' in \emph{2015 IEEE International Conference on
  Robotics and Automation (ICRA)}, May 2015, pp. 1795--1801.

\bibitem{dames2017detecting}
P.~Dames, P.~Tokekar, and V.~Kumar, ``Detecting, localizing, and tracking an
  unknown number of moving targets using a team of mobile robots,'' \emph{The
  International Journal of Robotics Research}, vol.~36, no. 13-14, pp.
  1540--1553, 2017.

\bibitem{liu2017kalman}
Q.~Liu, Z.~Wang, X.~He, and D.~Zhou, ``On kalman-consensus filtering with
  random link failures over sensor networks,'' \emph{IEEE Transactions on
  Automatic Control}, vol.~63, no.~8, pp. 2701--2708, 2017.

\bibitem{Pasqualetti2014}
F.~{Pasqualetti}, S.~{Zampieri}, and F.~{Bullo}, ``Controllability metrics,
  limitations and algorithms for complex networks,'' in \emph{2014 American
  Control Conference}, June 2014, pp. 3287--3292.

\bibitem{leitold2017controllability}
D.~Leitold, {\'A}.~Vathy-Fogarassy, and J.~Abonyi, ``Controllability and
  observability in complex networks--the effect of connection types,''
  \emph{Scientific reports}, vol.~7, no.~1, p. 151, 2017.

\bibitem{Ramachandran2017}
R.~K. {Ramachandran} and S.~{Berman}, ``The effect of communication topology on
  scalar field estimation by large networks with partially accessible
  measurements,'' in \emph{2017 American Control Conference (ACC)}, May 2017,
  pp. 3886--3893.

\bibitem{ramachandran2019resilience}
\BIBentryALTinterwordspacing
R.~K. Ramachandran, J.~A. Preiss, and G.~S. Sukhatme, ``Resilience by
  reconfiguration: Exploiting heterogeneity in robot teams,'' in \emph{IEEE/RSJ
  Int'l. Conf. on Intelligent Robots and Systems (IROS)}, 2019, to appear.
  [Online]. Available: \url{https://arxiv.org/abs/1903.04856}
\BIBentrySTDinterwordspacing

\bibitem{sharma2010sensor}
A.~B. Sharma, L.~Golubchik, and R.~Govindan, ``Sensor faults: Detection methods
  and prevalence in real-world datasets,'' \emph{ACM Transactions on Sensor
  Networks (TOSN)}, vol.~6, no.~3, p.~23, 2010.

\bibitem{arosh2015fitness}
S.~Arosh, S.~Nayak, S.~Duttagupta, \emph{et~al.}, ``Fitness function based
  sensor degradation estimation using h infinity filter,'' \emph{Procedia
  Computer Science}, vol.~58, pp. 172--177, 2015.

\bibitem{jiang2006sensor}
L.~Jiang, D.~Djurdjanovic, J.~Ni, and J.~Lee, ``Sensor degradation detection in
  linear systems,'' in \emph{Engineering Asset Management}.\hskip 1em plus
  0.5em minus 0.4em\relax Springer, 2006, pp. 1252--1260.

\bibitem{Bonilla2017}
L.~{Guerrero-Bonilla}, A.~{Prorok}, and V.~{Kumar}, ``Formations for resilient
  robot teams,'' \emph{IEEE Robotics and Automation Letters}, vol.~2, no.~2,
  pp. 841--848, April 2017.

\bibitem{Horn:1985:MA:5509}
R.~A. Horn and C.~R. Johnson, Eds., \emph{Matrix Analysis}.\hskip 1em plus
  0.5em minus 0.4em\relax New York, NY, USA: Cambridge University Press, 1986.

\bibitem{FB-LNS}
\BIBentryALTinterwordspacing
F.~Bullo, \emph{Lectures on Network Systems}, 1st~ed.\hskip 1em plus 0.5em
  minus 0.4em\relax Kindle Direct Publishing, 2019, with contributions by J.
  Cortes, F. Dorfler, and S. Martinez. [Online]. Available:
  \url{http://motion.me.ucsb.edu/book-lns}
\BIBentrySTDinterwordspacing

\bibitem{olfati2007dkf}
Olfati-Saber, ``Distributed kalman filtering for sensor networks,'' in
  \emph{Decision and Control, 2007 46th IEEE Conference on}, 2007.

\bibitem{morbidi2011active}
F.~Morbidi and G.~L. Mariottini, ``On active target tracking and cooperative
  localization for multiple aerial vehicles,'' in \emph{2011 IEEE/RSJ
  International Conference on Intelligent Robots and Systems}.\hskip 1em plus
  0.5em minus 0.4em\relax IEEE, 2011, pp. 2229--2234.

\bibitem{battistelli2012consensus}
G.~Battistelli, L.~Chisci, G.~Mugnai, A.~Farina, and A.~Graziano,
  ``Consensus-based algorithms for distributed filtering,'' in \emph{2012 IEEE
  51st IEEE Conference on Decision and Control (CDC)}.\hskip 1em plus 0.5em
  minus 0.4em\relax IEEE, 2012, pp. 794--799.

\bibitem{rafiee2010optimal}
M.~Rafiee and A.~M. Bayen, ``Optimal network topology design in multi-agent
  systems for efficient average consensus,'' in \emph{49th IEEE Conference on
  Decision and Control}.\hskip 1em plus 0.5em minus 0.4em\relax IEEE, 2010, pp.
  3877--3883.

\bibitem{xue2014distributed}
D.~Xue, A.~Gusrialdi, and S.~Hirche, ``A distributed strategy for near-optimal
  network topology design,'' in \emph{21st International Symposium on
  Mathematical Theory of Networked and Systems (MTNS 2014)}, 2014.

\bibitem{LaValle:2006:PA:1213331}
S.~M. LaValle, \emph{Planning Algorithms}.\hskip 1em plus 0.5em minus
  0.4em\relax New York, NY, USA: Cambridge University Press, 2006.

\bibitem{crazyswarm}
J.~A. Preiss*, W.~H\"onig*, G.~S. Sukhatme, and N.~Ayanian, ``Crazyswarm: {A}
  large nano-quadcopter swarm,'' in \emph{{IEEE} International Conference on
  Robotics and Automation ({ICRA})}.\hskip 1em plus 0.5em minus 0.4em\relax
  {IEEE}, 2017, pp. 3299--3304.

\end{thebibliography}
